\documentclass[final,1p,times,number]{elsarticle}

\usepackage[numbers]{natbib}
\usepackage{amsmath}
\usepackage{graphicx}
\usepackage{notoccite}
\usepackage{pdflscape}
\usepackage{longtable}  
\usepackage{booktabs}
\usepackage{tabularx}
\usepackage{hyperref}
\usepackage{multirow}
\usepackage{txfonts}
\usepackage{multirow}
\usepackage{placeins}

\begin{document}



\title{SISA-Rec: A Semantically Integrated Sequential Recommender with Contrastive Alignment}                      



\author[1]{Soohan Abbasi}
\ead{soohan.abbasi@dsu.edu.pk}
 
\author[2]{Shahid Munir Shah}
\ead{dr.shahid@ghr.szabist.edu.pk}
 
\author[3]{Rafia Shaikh}
 
\author[4]{Mahmoud Aljawarneh}
\ead{ma_jawarneh@asu.edu.jo}
 
\affiliation[1]{organization={Department of Computer Science, DHA Suffa University},
    city={Karachi},
    state={Sindh},
    country={Pakistan}}
 
\affiliation[2]{organization={Department of Computer Science, SZABIST},
    country={Pakistan}}
 
\affiliation[3]{organization={National University of Computer and Emerging Sciences (FAST)},
    city={Karachi},
    state={Sindh},
    country={Pakistan}}
 
\affiliation[4]{organization={Department of Computer Science, Applied Science Private University},
    city={Amman},
    country={Jordan}}

\begin{abstract}
Recommendation systems help users recommend relevant items from a large collection of choices. Present work on transformer-based sequential recommendation learns user preferences from interaction logs, but it mostly focuses on item identifiers and doesn't fully use the semantic meaning of items. This limitation becomes a major challenge in sparse and cold-start scenarios where historical interaction data is limited. 
To solve this problem, we introduce SISA-Rec (Semantically Integrated Sequential Recommendation), a transformer-based framework that embeds semantic context directly into sequential modeling. Our approach fuses item ID embeddings with BERT-based text embeddings via a gated fusion module, injects semantic similarity into the self-attention mechanism, and leverages an attention-based aggregation module to construct comprehensive user representations. Finally, a joint learning objective which combines Bayesian Personalized Ranking (BPR) and contrastive alignment loss, aligns the underlying behavioral and semantic spaces.
Experiments were conducted on the two highly sparse Amazon Beauty and Amazon Toys \& Games datasets, both having 99.93\% sparsity. The results show that SISA-Rec outperforms state-of-the-art baseline models across all evaluation metrics. Compared with the BERT4Rec \cite{petrov2022systematic}, SISA-Rec improves HR@10 by 16.6\% and NDCG@10 by 10.3\% on Amazon Beauty, and HR@10 by 23.1\% and NDCG@10 by 17.9\% on Amazon Toys \& Games.
Cold-start analysis further shows that the proposed model achieves the largest improvements for users with limited interaction historical records. This showcases the value of semantic information when user behavior data is scarce. Overall, the results demonstrate that integrating semantic information into the attention mechanism leads to more accurate and reliable recommendations




\end{abstract}




\maketitle

\section{Introduction}

Recommender systems have become a fundamental part of how people engage with online platforms. With the continued growth of  digital content in  e-commerce, streaming, and social media, the objective of these systems is to identify a manageable set of relevant items from a large pool of available options for each user. Recent advances in deep learning have significantly improved recommendation performance by learning richer representations from user behavior. Among these, sequential recommendation has drawn a lot of attention, since treating a user's interaction history as an ordered sequence makes it possible to capture how preferences shift over time rather than assuming they stay fixed. Transformer-based architectures such as SASRec and BERT4Rec push this further through self-attention, which lets the model weigh relationships between items across an entire sequence and pick up the contextual dependencies that simpler models tend to miss.
Alongside these developments, there has been a steady effort to bring item content, or semantics, more directly into the recommendation process. AI Models like UniSRec and MoRec have shown that textual and modality-based item representations are valuable to include, with modality-based encoders matching or even improving upon identity-based models in several settings. Knowledge graph approaches such as DiffKG and Gformer take a different approach, using structured relational data to capture higher-order semantic connections between items. Overall, these studies suggest that an item's content and semantic characteristics, rather than only its co-occurrence patterns, provide useful signals for recommendation.
The difficulty is that these two strengths rarely come together in the same model. Sequential and transformer-based methods, for all their ability to model temporal patterns, share a common dependence on item identifier embeddings learned purely from interaction data, which leaves them with very little sense of what an item means at a content level. That dependence is not much of a problem on dense benchmarks, where co-occurrence patterns alone are rich enough to support good recommendations. In realistic settings, however, catalogs are large and interaction data is extremely sparse, and cold-start users with only a handful of interactions are the norm rather than the exception. Under these conditions the behavioral signal is simply too thin to carry the full load, and a model that understands nothing about item content has little to fall back on. Semantic-aware models have the opposite limitation: they capture item meaning reasonably well but tend to overlook how a user's preferences evolve across an interaction sequence.
A key observation across the literature is that semantic information, even when it is used, is almost always treated as a secondary signal. It gets processed through a separate pathway, fused with behavioral representations only after the sequential encoding is already finished, or attached as a plug-and-play module on top of an otherwise unchanged architecture. None of these designs lets semantic relationships influence the sequential modeling itself. In particular, there is no mechanism that fuses identity and semantic representations at the feature level inside the attention computation, which is where the actual sequence modeling takes place. The result is a persistent separation between behavioral modeling and semantic grounding, and that separation imposes a real ceiling on existing frameworks. The ceiling is most visible exactly where it matters most, in the sparse and cold-start conditions that behavioral signals alone cannot handle.
This work starts from that gap. We propose SISA-Rec, a Semantically Integrated Sequential recommender designed so that semantic representations stay active at every stage of encoding instead of being added on at the end. Item identifier embeddings and frozen BERT semantic embeddings are combined through a gated fusion module at the input layer, so that semantics shapes the item representation before any sequential modeling happens. A semantic similarity matrix is then injected directly into the self-attention score at each transformer layer, which allows semantically related items to influence attention regardless of whether they have ever co-occurred in the observed history. An attention-based aggregation module builds the user representation as a weighted sum over all positions in the sequence rather than relying on the final hidden state alone, and a joint training objective combining a Bayesian Personalized Ranking loss with a contrastive alignment loss keeps the behavioral and semantic user representations consistent with one another. The framework is evaluated on the Amazon Beauty and Amazon Toys and Games datasets, both of which exhibit a sparsity level of 99.93\%, in order to test exactly the data-scarce regime where semantic grounding is expected to matter most.
The main contributions of this work are summarized as follows:

\begin{itemize}
\item A gated fusion module that embeds item IDs and feeds projected BERT semantic representations into the input layer, learning per-dimension how much of each source to retain.

\item A modified self-attention mechanism in which a semantic similarity matrix is fed directly into the attention score computation at each transformer layer.

\item An attention-based preference aggregation module that produces a single user vector through a learned weighted sum over all positions in the sequence.

\item A joint training objective combining a BPR loss with a contrastive alignment loss, so that the behavioral and semantic user representations are explicitly aligned.

\item An architecture aimed at improving generalization in data-sparse and cold-start settings, where semantic signals supply relational information that does not depend on item co-occurrence frequency, validated empirically on the Amazon Beauty and Amazon Toys and Games datasets, both 99.93\% sparse.

The rest of the paper has been outlines as follows:
Section \ref{LR} presents the review of the existing literature related to the sequential recommendation systems. 

\end{itemize}

\section{Literature Review}
\label{LR}

This section covers recent work on transformer-based and semantic-aware recommender systems. The studies brought together here investigate different ways of combining user interaction sequences with semantic item representations to improve how well recommendations serve users. The table in the next section summarizes the key aspects of each reviewed work.


Petrov and Macdonald \cite{petrov2022systematic} systematically reviewed the reproducibility assessement of BERT4Rec for sequential recommendation. BERT4Rec uses masked self-attention to learn bidirectional user preferences from interaction sequences and generally outperforms unidirectional models. However, the performance of the model notably depends on hyperparameter tuning and requires intensive computational resources.

Petrov \cite{petrov2024effective} evaluated the performance under scalability and cost limitations of transformer-based sequential recommender systems. Whereas applying NLP transformers to predict user interactions yields strong benchmark results, the computational expense is substantial and increases significantly with larger item inventories. For example, training BERT4Rec on the relatively small MovieLens-1M dataset takes up to 20 GPU hours. The study concludes that benchmark accuracy is of little value if models cannot scale to real-world, million-item catalogs, arguing that current training and inference strategies have hit a limit in their refinement.


Zhou et al. \cite{10.1145/3485447.3512111} proposed FMLP-Rec, a filter-enhanced MLP-based sequential recommendation model that replaces self attention by a filter that transforms input representations into the frequency domain via the Fast Fourier Transform, in order to denoise them, and transforms them back via the inverse transform. The authors tested on eight datasets, and got relatively stable results, with their all-MLP architecture and learnable filters outperforming several deep learning baselines. The downside of the model is it relies solely on item ID embeddings and does not incorporate semantic item information.

Tikhonovich et al \cite{tikhonovich2025esasrec} introduced eSASRec, a new, improved transformer-based sequential recommendation (SASRec) architecture to enhance the performance of current SASRec models. The study examines a series of discrete upgrades that can be made to transformer-based recommendation systems by breaking down the transformer layer structure, training, loss, and negative sampling strategies. The eSASRec model brings together SASRec's sequence training objective, LiGR transformer, and Sampled Softmax loss. Testing spanned ML-20M, Kion, and BeerAdvocate datasets using NDCG@10, Recall@10, and coverage judge quality. The model showed up to 23\% better accuracy than the state-of-the-art methods of the standard academic datasets. When the evaluation conditions were made more realistic, however, the margin over industrial models like HSTU shrank considerably, which is itself an important finding about how much the choice of evaluation protocol can quietly tilt what a study appears to demonstrate.

Hou et al. \cite{10.1145/3534678.3539381} proposed UniSRec, a universal sequential recommendation framework that uses item textual descriptions instead of item IDs to improve cross-domain transferability and relove the  problem when the item catalogues involved have no items in common. UniSRec includes a lightweight architecture combining parametric whitening, a mixture-of-experts adaptor, and learnable gating for domain fusion. To build adaptable sequence representations, it relies on two contrastive pre-training tasks using multi-domain negative sampling. The results show this framework gives strong cross-platform performance on real-world datasets without requiring heavy re-training. Despite this, semantic information is processed separately from the attention mechanism, limiting its influence on sequential modeling.


Yuan et al. \cite{10.1145/3539618.3591932} reconsidered the debate among identity-based and modality-based recommendation approaches using text and visual representations. The central question driving the study is whether a purely modality-based model can match or surpass a pure identity-based model when the item identifier embedding is replaced with a strong modality encoder such as BERT or Vision Transformer. Their results demonstrated that modality-based methods can match or outperform identity-based approaches, particularly when semantic features are incorporated. Despite these findings, the study does not provide a mechanism for dynamically fusing identity and semantic representations at the feature level within the attention computation itself, which is precisely the direction taken in the present work through the proposed gated fusion module.

Qiu et al. \cite{10.1145/3488560.3498433} proposed DuoRec, a contrastive learning framework designed to tackle representation degeneration problem in sequential recommendation.The core issue is that existing sequential models produce item embeddings that form a narrow geometric cone. This causes rare items to cluster in the embedding space, making them seem more similar than they really are. DuoRec addressed this using a uniformity-based contrastive regularization objective to distribute sequence representations more evenly. Rather than standard augmentations like cropping or masking, it used model-level dropout to generate semantically consistent training samples. It also provide a novel sampling strategy, treating sequences with the same target item as hard positives. Five benchmark datasets were used for testing, and DuoRec came out ahead of standard sequential recommendation baselines across all of them. Nevertheless, it relies entirely on item ID embeddings and lacks semantic item representations.

Zhai et al. \cite{li2023text} proposed a transformer-based sequential recommendation framework centered on the textual information that items carry. The model treats this as a language task, leveraging pre-trained models to pull semantics from item descriptions. These representations then feed into sequential recommendation models, giving them a better basis for understanding how items relate to one another semantically within a user's interaction history. Experiments were conducted on Amazon and MovieLens datasets, with NDCG and Hit Ratio used to evaluate ranking quality. The text-enriched models performed better than the traditional ID-based models in all experiments. The downside to this approach is that the framework is highly reliant on descriptive text to establish a similarity score, and in environments where text-enriched information is limited or unavailable, the
performance is significantly impaired.

Li et al. \cite{li2023e4srec} proposed E4SRec, a framework connecting large language models to conventional sequential recommender systems while keeping things efficient and easy to extend. The problem with most existing large language model-based recommendation work is that it turns everything into an open-ended generation task, which means items need rich semantic descriptions, the outputs often miss the valid candidate range, and the whole thing runs slowly. E4SRec feeds item identifier sequences directly into the model and keeps outputs anchored to valid candidate lists, generating a full ranking in a single forward pass with just a handful of pluggable parameters trained per dataset and the large language model left entirely frozen. Amazon Beauty, Sports, Toys, and Yelp benchmarks all pointed to the approach working well in terms of both quality and speed. What the model does not do is build an explicit bridge between semantic item representations and sequential user behavior inside the attention layer, so whatever semantic understanding is present comes from what the large language model already learned during pre-training rather than from any deliberate alignment mechanism.

Wang et al. \cite{wang2023knowledge} proposed a knowledge-aware collaborative filtering framework that brings a pre-trained language model to strengthen personalized review-based rating prediction. Review text and external knowledge representations both feed into the model, giving it more material to work with when mapping semantic connections between users and items. Passing reviews through a pre-trained language model produces a depth in user-item representations that interaction-only collaborative filtering approaches typically cannot reach. Results on review-based benchmark datasets showed stronger rating prediction than conventional collaborative filtering baselines. That said, stacking knowledge representations and a language model together across multiple stages is not simple to manage, and the added complexity does show up in the form of higher training costs and heavier computational load.

Jiang et al. \cite{jiang2024diffkg} proposed a new DiffKG that introduces a diffusion model to knowledge graph-based collaborative filtering. The diffusion model is essentially taking knowledge graph relations, breaking them down gradually, and then putting them back together, and then repeating the same thing (this process gets rid of the less useful and noisy edges). The framework also includes a knowledge graph convolution module, which uses user-item interaction information combined with knowledge graph information to help better understand user preferences. The model was tested on three datasets: Last-FM, MIND, and Alibaba-iFashion using Recall@20 and NDCG@20 as the criteria.
The model came out on top across all three, putting up Recall@20 of 0.0980 and NDCG@20 of 0.0911 on Last-FM, Recall@20 of 0.0615 and NDCG@20 of 0.0389 on MIND, and Recall@20 of 0.1234 and NDCG@20 of 0.0773 on Alibaba-iFashion. All in all, this shows that it's probably worth the effort to tweak knowledge graph representations to get better recommendations.

Fan et al. \cite{li2023graph} introduced a Graph Transformer recommender. The simple premise is that a combination of graph neural networks and transformer attention is used to boost the ability to work with relationships in the user-item graph. The self-attention mechanism is applied to graph data to learn representations of nodes that have semantic meaning, and higher-order user-item relationships are maintained. The reason for combining the graph message passing with transformer attention is that graph-based recommenders alone have problems with some types of relationships, and the transformer attention helps bridge that. On a couple of benchmark data sets, the model was able to better capture long-range dependencies and complex user-item relationships than typical graph neural network models, which translated to better recommendations. The obvious tradeoff is complexity. Combining graph propagation with transformer attention isn't cheap, and it shows in that the training and inference times are higher.

Xu et al. \cite{cui2025sagerec} developed SAGERec, a recommendation model that uses semantic information. The idea is that semantic representations of items, together with user-item interactions, should allow the model to better understand user preferences and item characteristics than just considering interactions. This was confirmed by benchmarking tests using Hit Ratio and NDCG, where SAGERec outperformed traditional collaborative filtering methods that only use interaction data. But the thing to note is how the semantic information is used within the model. It is a plug-and-play extension of the model, rather than being integrated directly with the sequential attention mechanism. As a result, semantic representations cannot react to a user's changing interests over time given their interaction history, and it limits the extent to which the semantic component can affect the final model.

\renewcommand{\arraystretch}{1.35}

\begin{longtable}{|p{1.6cm}| p{1.6cm} |p{1.6cm} |p{2cm}| p{2.8cm} |p{3cm}|}
\caption{Comparative Analysis of Sequential and Semantic-Aware Recommendation Approaches}
\label{tab:comparison} \\
\hline
\textbf{Reference} & \textbf{Model} & \textbf{Item Representation} & \textbf{Semantic Integration} & \textbf{Sequential Modeling} & \textbf{Results} \\
\hline

\cite{petrov2022systematic}  & BERT4Rec  & Item ID embeddings & None & Bidirectional Transformer & Recall@10 = 0.6975, NDCG@10 = 0.4751 \\
\hline
\cite{petrov2024effective} & Efficient Transformer CF framework & Item ID embeddings & None & Transformer self-attention & Numerical benchmarks not reported in extended abstract \\
\hline
\cite{10.1145/3485447.3512111} & FMLP-Rec & Item ID embeddings & None & All-MLP with frequency-domain filtering & NDCG@10 = 0.0561, Recall@10 = 0.0931 \\
\hline
\cite{tikhonovich2025esasrec} & eSASRec & Item ID embeddings & None & Transformer sequential modeling & NDCG@10 = 0.1563 \\
\hline
\cite{10.1145/3534678.3539381} & UniSRec & Text embeddings & Cross-domain semantic transfer via MoE adaptor & Transformer sequential modeling & NDCG@10 = 0.1379, Recall@10 = 0.2316 \\
\hline
\cite{10.1145/3539618.3591932} & MoRec & Modality-based embeddings (BERT/ViT) & End-to-end modality encoder & Transformer sequential modelingn & HR@10 = 0.0623 (competitive with IDRec \\
\hline
\cite{10.1145/3488560.3498433} & DuoRec & Item ID embeddings & None & Transformer with contrastive regularization & NDCG@10 = 0.0437, Recall@10 = 0.0835 \\
\hline
\cite{li2023text} & Recformer & Text embeddings & Text semantic representations & Transformer sequential modeling & NDCG@10 = 0.1027, Recall@10 = 0.1448 \\
\hline
\cite{li2023e4srec} & E4SRec & Item ID embeddings + LLM & Implicit via frozen LLM & LLM-based sequential generation & NDCG@10 = 0.0892, Recall@10 = 0.1534 \\
\hline
\cite{wang2023knowledge} & KCF-PLM & Text + Knowledge Graph embeddings & Dual semantic alignment & Limited sequential modeling & MSE = 0.8740 \\
\hline
\cite{jiang2024diffkg} & DiffKG & Knowledge graph embeddings & Diffusion-based semantic propagation & Order-invariant (no sequential modeling) & Recall@20 = 0.0980, NDCG@20 = 0.0911 \\
\hline
\cite{li2023graph} & Gformer & Graph node embeddings & Graph relational semantics & Transformer-based graph attention & Recall@20 = 0.0878, NDCG@20 = 0.0442 \\
\hline
\cite{cui2025sagerec} & SAGERec & Semantic item embeddings & Semantic representation fusion & Sequential recommendation modeling & NDCG@10 = 0.1288, Recall@10 = 0.2680 \\
\hline
\end{longtable}

In table \ref{tab:comparison}, the reviewed papers are synthesized with an emphasis on the comparative analysis of approaches based on transformers and semantics. For each paper, several aspects are analyzed, namely the type of architecture used, item encoding, semantics incorporation, sequential modeling technique, and results achieved. Through the synthesis, similarities and differences between methods are pointed out, focusing on the way semantics were incorporated and interaction ordering was addressed. Papers come from different architecture backgrounds, namely sequential transformer architectures, contrastive learning, multimodal encoder architectures, knowledge graph-based techniques, and language models. The evaluation metrics presented throughout the studies include, among others, Recall, NDCG, HR, and MSE, all established standards in the realm of recommendation studies.

\subsection{Problem Identification}

Recommendation systems research has moved forward quite a bit over the years, and most of that progress traces back to advances in deep learning and representation learning. The studies covered here are methodologically varied, drawing from collaborative filtering, sequential recommendation, transformer-based architectures, contrastive learning, modality-based encoders, and knowledge graph-driven systems. Each has contributed something useful, either by improving how systems handle user-item interactions or by getting better at pulling preference patterns out of behavioral data. Sequential models like BERT4Rec, eSASRec, and FMLP-Rec handle temporal dependencies reasonably well by treating interaction histories as ordered sequences. Transformer-based methods strengthen this further through self-attention mechanisms that span entire sequences and pick up contextual relationships between items. That said, all of these models share a common dependency on item identifier embeddings derived from interaction data, and that leaves them with very little understanding of what items actually mean at a content level.

There has also been a parallel effort to bring semantic item knowledge more directly into recommendation models. UniSRec and MoRec have shown that incorporating textual item representations into recommendation pipelines is worthwhile, with modality-based encoders matching or outperforming identity-based models in several settings. DiffKG and Gformer take a different route, using structured relational data from knowledge graphs to capture higher-order semantic connections. That said, semantic information in most of these methods still ends up being treated as secondary, processed through a separate pathway, or combined with behavioral representations only after sequential encoding is already complete.

DuoRec makes a fair point about representation quality, showing that regularizing embedding distributions produces real gains in sequential recommendation. E4SRec reflects a growing recognition that large language models can bring richer semantic understanding into recommendations, and the results support this. That said, neither work directly tackles the alignment problem between semantic item characteristics and the sequential attention mechanism.

Looking across everything, one gap remains consistently unaddressed. Sequential models handle behavioral patterns well but have no semantic grounding for the items involved. Semantic-aware models capture item meaning but largely overlook how user preferences evolve. That separation creates a real ceiling for existing frameworks, and the problem is sharpest in data-sparse conditions and cold-start scenarios where behavioral signals alone cannot carry the full load. The present work starts from this gap, aiming to build a system where semantic representations are active at every stage of encoding rather than being added on at the end, with empirical validation conducted on the Amazon Beauty and Amazon Toys \& Games datasets, both of which exhibit a sparsity level of $99.93\%$.

\subsection{Our Contributions}

SISA-Rec is a new recommender system based on Top-N collaborative filtering with a semantically integrated self-attention network. The summary of the contribution of this work is as follows:
\begin{itemize}
\item A gated fusion module, which embeds item ID and feeds BERT semantic representation into the input layer.
\item An attention mechanism with a modified self-attention mechanism, where a semantic similarity matrix is directly fed into the attention score calculation at each transformer layer.
\item An attention-based preference aggregation module to generate a single user vector by taking a weighted sum over all positions in the sequence based on a learned weight.
\item A joint training objective based on both the BPR loss and a contrastive alignment loss to ensure that the behavioral and semantic user representations are aligned.
\item An architectural design inspired by the need to improve generalization in data-sparse settings and cold-start scenarios, where relational information is provided by semantic signals that do not depend on the co-occurrence frequency of data items, and validated empirically on the Amazon Beauty and Amazon Toys and Games data sets, both $99.93\%$ sparse.
\end{itemize}

\section{Methodology}

\begin{figure}[ht]
    \centering
    \includegraphics[width=1\textwidth]{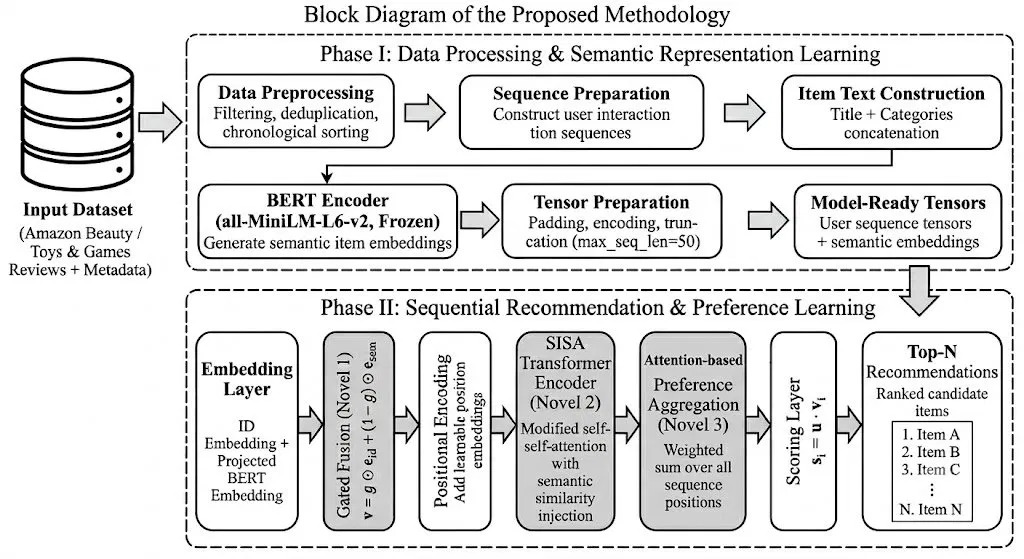}
    \caption{Block diagram of the proposed methodology.}
    \label{fig:fig-1}
\end{figure}

The methodology used in this research is based on the block diagram given in Fig.~\ref{fig:fig-1}

As shown in Fig.~\ref{fig:fig-1}, the proposed design adopts a dual-stage structure that combines data processing with sequential recommendation and preference learning. The input to the system consists of two Amazon product review datasets, namely Amazon Beauty and Amazon Toys \& Games, each comprising raw user-item interaction records and item metadata including product titles and category labels.

During the first stage, the input dataset is processed through a structured pipeline. Raw interaction and metadata records are first subjected to data preprocessing, which involves filtering, cleaning, and integration of the two data sources. The cleaned data is then used to construct chronologically ordered user interaction sequences. These sequences are subsequently transformed into fixed-length tensor representations through padding, encoding, and conversion operations, yielding model-ready tensors that serve as input to the second stage.

During the second stage, the model-ready tensors are fed into the sequential recommendation pipeline. Each item in the user sequence is first mapped to a dense representation through an embedding module that learns item representations from both identifier-based and semantic sources. These embeddings are then passed into a transformer encoder that models sequential patterns and captures long-range dependencies across the user interaction history. An attention-based aggregation mechanism subsequently processes the encoded sequence to capture user preferences by computing a weighted combination of all encoded positions. A ranking module finally uses the resulting user preference vector to score candidate items, and the top-ranked items are returned as the final Top-N recommendations.

\subsection{Dataset acquisition}

The experiments in this paper are conducted on two publicly available product-review datasets drawn from the Amazon Review Data collection~\cite{mcauley2015image}, namely \textit{Amazon Beauty} (primary dataset) and \textit{Amazon Toys \& Games} (secondary dataset). Both datasets record explicit user--item interactions in the form of product ratings accompanied by timestamps, enabling chronological ordering of each user's interaction history. In addition to interaction records, every item is associated with rich textual metadata, including product title, description, and category information, which is leveraged for semantic embedding learning in the proposed SISA-Rec framework.

Both datasets exhibit high sparsity ($\approx$99.93\%), making them substantially more challenging than commonly used dense benchmarks such as MovieLens-1M(sparsity $\approx$95.53\%). This property makes them well-suited for evaluating the generalization ability of sequential recommenders in realistic, data-scarce conditions, and for validating the hypothesis that semantic integration provides a
meaningful compensatory signal when collaborative filtering signals are limited.

Following common preprocessing practice in sequential recommendation~\cite{kang2018self}, only users with at least five interactions are retained. The dataset is partitioned into training, validation, and test sets using a leave-one-out chronological strategy: the most recent interaction of each user is held out for testing, the second most
recent for validation, and all remaining interactions form the training set. This ensures that every prediction is made strictly from past observations, preserving temporal causality throughout evaluation.

The summary statistics for both datasets after preprocessing are presented in Table~\ref{tab:dataset_stats}.

\begin{table}[ht]
\centering
\caption{Dataset Statistics after Preprocessing}
\label{tab:dataset_stats}
\begin{tabular}{lcc}
\toprule
\textbf{Parameter}          & \textbf{Amazon Beauty} & \textbf{Amazon Toys \& Games} \\
\midrule
Total Users                 & 22{,}363               & 19{,}412                      \\
Total Items                 & 12{,}101               & 11{,}924                      \\
Total Interactions          & 198{,}502              & 167{,}597                     \\
Average Sequence Length     & 8.88                   & 8.63                          \\
Dataset Sparsity            & 99.93\%                & 99.93\%                       \\
\bottomrule
\end{tabular}
\end{table}

For semantic representation learning, the textual description of each item $i \in \mathcal{I}$ is formed by concatenating its title and primary category label, following a similar methodology to~\cite{10.1145/3534678.3539381}:
\begin{equation}
    x_i = \mathrm{title}_i \oplus \mathrm{category}_i
    \label{eq:text_concat}
\end{equation}
where $\oplus$ denotes string concatenation. This produces a natural-language representation for every item that is subsequently encoded by the frozen BERT encoder to obtain semantic embeddings, as described in Section~\ref{sec:embedding}.

\subsection{Data Preprocessing}
\begin{figure}[ht]
\centering
\includegraphics[width=0.6\linewidth]{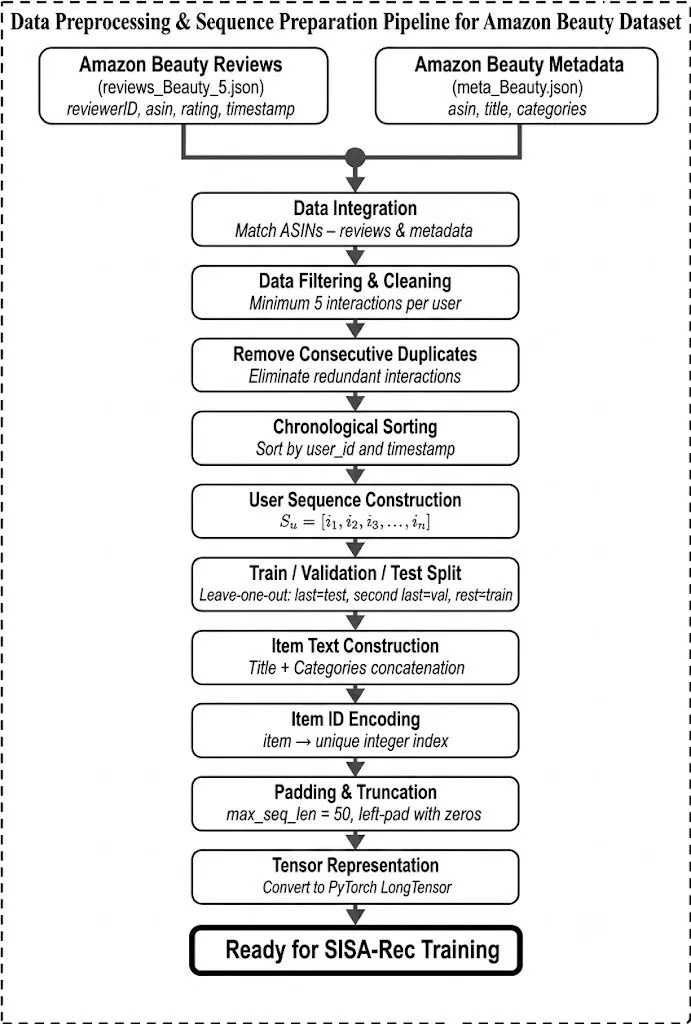}
\caption{Data preprocessing and sequence preparation pipeline for Amazon Beauty and Amazon Toys \& Games datasets.}
\label{fig:pipeline}
\end{figure}

The raw datasets of Amazon Beauty and Amazon Toys \& Games are processed through a structured pipeline to guarantee data quality and suitability for sequential modeling. The rating file and the item metadata file are merged on the common item identifier, resulting in a unified representation where each user interaction is paired with its corresponding item-level information. A minimum interaction threshold is applied to retain only users with an adequate interaction history, following common preprocessing practices in sequential recommendation~\cite{kang2018self}. Formally, a user $u \in \mathcal{U}$ is retained if and only if:
\begin{equation}
|R_u| \geq \theta
\label{eq:filter}
\end{equation}
where $R_u$ is the interaction set for user $u$, $\mathcal{U}$ is the set of all users, and $\theta = 5$. The repeated consecutive transactions are then removed to prevent redundancy, following common sequence-cleaning techniques~\cite{kang2018self}. Mathematically, for a user sequence $S_u = [i_1, i_2, \ldots, i_n]$, any item $i_t$ such that:
\begin{equation}
i_t = i_{t-1}
\label{eq:dedup}
\end{equation}
is deleted, resulting in a deduplicated series of unique item changes. The dataset is then ordered by the reviewer's ID and timestamp to ensure the correct chronological order of the transactions, which is critical for sequential models. The complete preprocessing pipeline is illustrated in Fig~\ref{fig:pipeline}

\subsection{Sequence Preparation}
After data preprocessing, the interaction logs are ordered as chronological user sequences for modeling purposes~\cite{kang2018self}. For each user $u \in \mathcal{U}$, interactions are collected and sorted according to time stamp, forming a chronological sequence of interaction activities:

\begin{equation}
    S_u = [i_1, i_2, i_3, \ldots, i_n]
    \label{eq:user_sequence}
\end{equation}

where $i_t \in \mathcal{I}$  refers to the interaction on item $t$, and $n$ refers to the total 
number of interactions carried out by the user $u$. item order in $S_u$ is maintained to capture the chronology in which the user performed interactions, guaranteeing that any potential dependencies on sequential interactions are captured in the process. To ensure chronological consistency between train, validate, and test sets, a temporal leave-one-out split method is employed as suggested in~\cite{he2017neural}. In particular, considering a user sequence $S_u$ of size $n$:

\begin{equation}
    S_u^{train} = [i_1, i_2, \ldots, i_{n-2}], \quad 
    S_u^{val} = [i_{n-1}], \quad 
    S_u^{test} = [i_n]
    \label{eq:data_split}
\end{equation}

The last interaction $i_n$ is used for testing, while the next-to-last interaction $i_{n-1}$ is used for validation, whereas the rest of the interactions form the training dataset. The above process guarantees that predictions of future interactions are made based only on past behavioral signals, thus maintaining temporal causality.

For semantic representation learning purposes, the concatenation of the title and category features of each individual item $i \in \mathcal{I}$ forms a textual representation, adopting a similar methodology of constructing item texts as in~\cite{10.1145/3534678.3539381}. Namely,

\begin{equation}
    x_i = \text{title}_i \oplus \text{category}_i
    \label{eq:text_concat2}
\end{equation}
where $\oplus$ is the operator for string concatenation. This operation translates the structured metadata to a natural language representation, enabling the transformer-based encoder to learn not only lexical but also contextual information among item attributes. The textual representation $x_i$ is used as input for the BERT encoder in the next step of semantic embedding learning.

\subsection{ Tensor Preparation}
For processing in neural network models, the created user sequence is converted into numeric tensors with a fixed length. Each item $i \in \mathcal{I}$ is initially encoded as a unique integer index using an encoder function, as is conventional in item indexing for sequential recommendation tasks~\cite{kang2018self}. The encoding 
function $f : \mathcal{I} \rightarrow \mathbb{Z}^+$ can be expressed as:

\begin{equation}
    f(i) = k, \quad k \in \{1, 2, \ldots, |\mathcal{I}|\}
    \label{eq:id_encoding}
\end{equation}

where $|\mathcal{I}|$ indicates the total number of distinct items available in the database. Index $0$ is solely allocated to pad positions in order to enable the model to differentiate between actual item engagements and padded positions.

In order to address the problem of varying sequence lengths, every user sequence is first turned into a fixed-length sequence by means of padding and truncation using Kang and McAuley~\cite{kang2018self}'s method of preparing user sequences. For a user sequence $S_u$ with length $k$, the padded sequence $\tilde{S}_u$ is: 

\begin{equation}
    \tilde{S}_u = 
    \begin{cases} 
    [0, \ldots, 0, i_1, \ldots, i_k] & \text{if } k < L \\
    [i_{n-L+1}, \ldots, i_n] & \text{if } n > L
    \end{cases}
    \label{eq:padding}
\end{equation}

for which we have assumed $L = 50$ as the length limit in this work. Shorter interaction sequences are filled from the left using zeros, thereby maintaining the recent interaction’s recency at the right end, while longer sequences are truncated, keeping the last $L$ interactions only. This particular method of truncation follows from our observation that the prediction of future preferences becomes easier with recent interactions rather than distant past interactions.

These zero-padded sequences are then encoded as PyTorch \texttt{LongTensor}, which will be used as input to the embedding layer of SISA-Rec.

\subsection{Embedding}
\label{sec:embedding}
Embedding converts the item indices generated from the tensor construction phase to continuous dense vectors compatible with neural architectures. In our SISA-Rec architecture, an embedding vector is constructed for each item $i \in \mathcal{I}$, where there exist two types of embeddings per item – an item ID embedding for collaborative filtering and an item semantic embedding for capturing the semantics.

\subsubsection{Identifier-Based Embedding}
Each item identifier is associated with a dense vector representation using a learnable embedding matrix, consistent with the identifier-based item embedding model proposed by Kang and McAuley~\cite{kang2018self}. Mathematically, the identifier-based embedding for item $i$ can be represented as follows:

\begin{equation}
    e^i_{id} = E_{id}[f(i)]
    \label{eq:id_embedding}
\end{equation}

where $E_{id} \in \mathbb{R}^{|\mathcal{I}| \times d}$ is a learnable embedding matrix, $d$ refers to the number of dimensions for the embeddings, and $f(i)$ refers to the integer identifier associated with the item $i$. The embeddings $E_{id}$ are randomly initialized and updated in the course of training using backpropagation.

\subsubsection{Semantic Embedding}

In order to leverage item semantic information, the textual embedding $x_i$ is computed by the BERT encoder using the item title and category information ~\cite{devlin2019bert}. Mathematically, this can be expressed as follows:

\begin{equation}
    e^i_{sem} = \text{BERT}(x_i) \in \mathbb{R}^{768}
    \label{eq:bert_embedding}
\end{equation}

Weights in the BERT model encoder remain frozen during training, since the pre-trained representation has already encoded complex lexical and contextual semantic relations that can be transferred from one domain to another~\cite{devlin2019bert}. To match the size of semantic embeddings and the size of the identifier-based embeddings, a dimensionality alignment layer using the weight matrix $W_p \in \mathbb{R}^{768 \times d}$ is applied~\cite{10.1145/3534678.3539381}

\begin{equation}
    e^i_{sem} = W_p \cdot \text{BERT}(x_i)
    \label{eq:projection}
\end{equation}

where the dimensionally projected semantic embedding $e^i_{sem} \in \mathbb{R}^d$ is the same as the identifier-based embedding $e^i_{id}$. Then, the two embeddings are fed to the gated fusion module to be merged.

\subsection{Gated Fusion}
The gated fusion module is the first new contribution of the SISA-Rec model. Unlike existing methods that either rely only on identifier-based embeddings or use semantic and semantic information as an additional, post-encoder module, the presented gated fusion module combines identifier-based and semantic representations at the very beginning of the encoding process, before applying sequential modeling. This guarantees that semantic information is involved in the construction of item representations that are subsequently encoded by the transformer encoder.

Given the identifier-based and projected semantic embeddings $e^i_{id} \in \mathbb{R}^d$ and $e^i_{sem} \in \mathbb{R}^d$ for each item $i \in \mathcal{I}$, we learn a gate vector $g \in [0, 1]^d$ 
is computed as proposed in this work:

\begin{equation}
    g = \sigma\left(W_g \cdot [e^i_{id} ; e^i_{sem}] 
    + b_g\right)
    \label{eq:gate}
\end{equation}

with $W_g \in \mathbb{R}^{d \times 2d}$ being a learnable weight matrix, $b_g \in \mathbb{R}^d$ being a bias vector, $[\cdot ; \cdot]$ being the vector concatenation operator, and $\sigma(\cdot)$ being the sigmoid function. The gate vector $g$ is in the range $[0, 1]^d$, which allows for individual control over the importance of each element of the identifier-based and semantic embeddings.

Finally, the item representation $v_i$ is calculated as a weighted sum of the two embeddings, as we propose here:

\begin{equation}
    v_i = g \odot e^i_{id} + (1 - g) \odot e^i_{sem}
    \label{eq:fused_representation}
\end{equation}

where $\odot$ is element-wise multiplication. The gate $g$ adaptively controls the amount of 
identifier-based and semantic information each dimension of the fused representation contains. When 
$g \rightarrow 1$, the fused representation is close to the identifier-based embedding, and if $g \rightarrow 0$,  the fused representation is close to the semantic embedding. This adaptive mechanism enables the model to learn the degree of fusion that optimally balances the information in the identifier-based embedding with the semantic embedding based on the data, rather than relying on a hand-tuned fusion weight.

The embedded representation $v_i \in \mathbb{R}^d$ is then fed to the positional encoding layer.

\subsection{Positional Encoding}
After gated fusion, positional encoding is used to encode the order of items in the sequence 
item representations. The transformer model is permutation-invariant by design, and does not contain a position encoding in its standard design, so positional encoding is important to allow the model to learn the temporal order of items in the user sequence. For a fused item representation $v_t \in \mathbb{R}^d$ at a position $t$ in the sequence, we add a learnable positional embedding $p_t \in \mathbb{R}^d$ to obtain the position-aware embedding $x_t$ as in Kang and McAuley~\cite{kang2018self}:

\begin{equation}
    x_t = v_t + p_t
    \label{eq:positional_encoding}
\end{equation}

where $p_t \in \mathbb{R}^d$ is a vector from a learnable matrix of positional embeddings $P \in \mathbb{R}^{L \times d}$, and $L$ is the maximum sequence length. Compared to fixed sinusoidal position embeddings~\cite{ashish2017attention}, learnable positional embeddings can adaptively learn the position-specific information for the recommendation task. The position-aware sequence $X = [x_1, x_2, \ldots, x_L] \in \mathbb{R}^{L \times d}$ is then fed into 
to the SISA transformer encoder.

\subsection{SISA Transformer Encoder}

\begin{figure*}[ht]
\centering
\includegraphics[width=\linewidth]{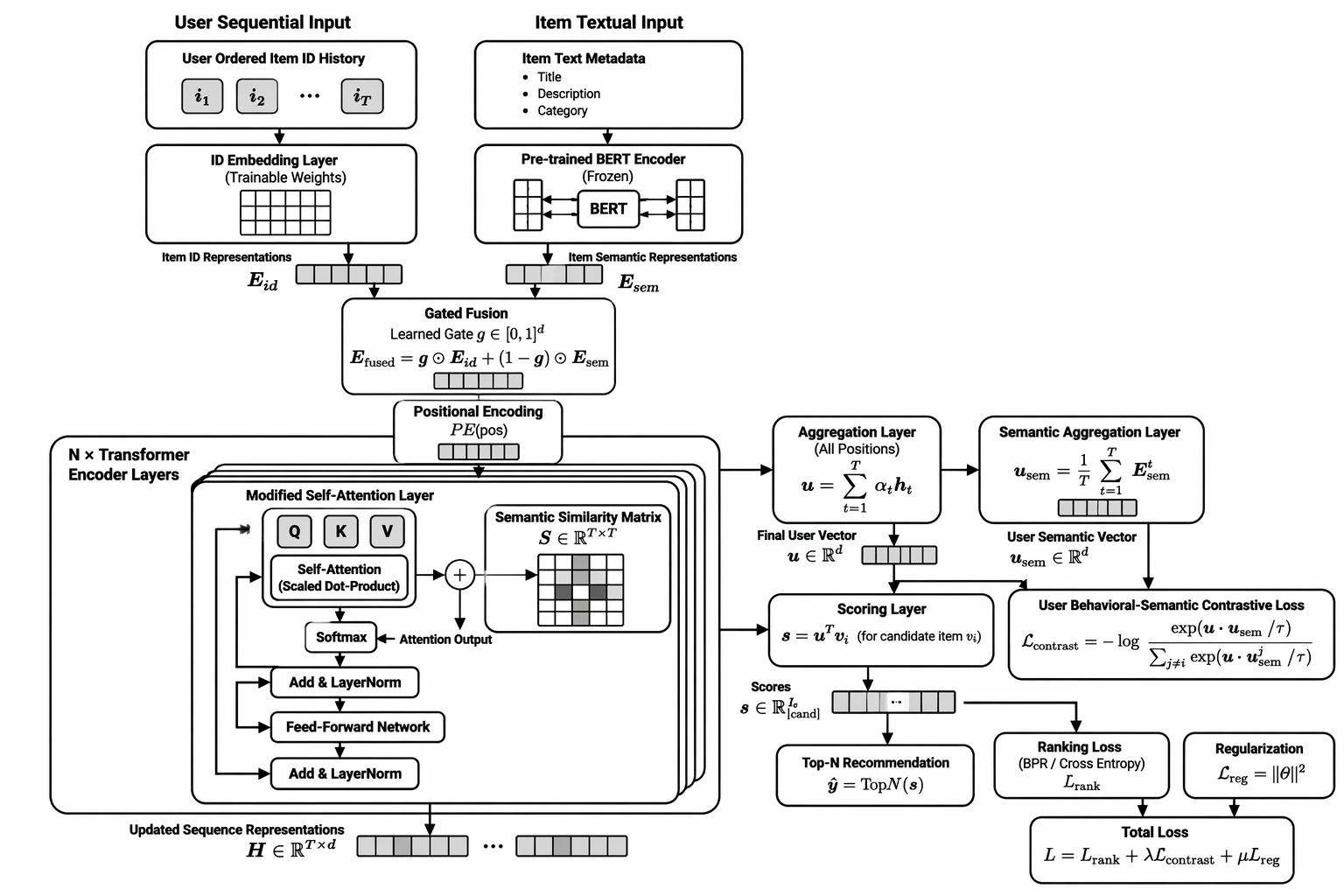}
\caption{Architecture of the proposed SISA-Rec model.}
\label{fig:architecture}
\end{figure*}

The SISA transformer encoder is the key sequence modeling block of the proposed architecture. Fig.~\ref{fig:architecture} shows the detailed architecture of the proposed SISA-Rec model.
It takes a sequence $\mathbf{X} \in \mathbb{R}^{L \times d}$ with positional information through $N$ transformer layers, which include a modified self-attention mechanism, a feed-forward network, layer normalization, and residual connections. The main innovation of the SISA encoder is to extend the self-attention mechanism to include semantic similarity signals as part of the attention computation, so that semantic relationships among items can drive the sequential modeling.

\subsubsection{Semantic Similarity Matrix}
Before the attention is computed, a semantic similarity matrix $\mathbf{S} \in \mathbb{R}^{T \times T}$ is computed from semantic embeddings of items in the sequence, provided by BERT The cosine similarity~\cite{manning2008introduction} between the semantic embeddings of the items at positions $i$ and $j$ is represented by $S[i, j]$, as we propose:

\begin{equation}
    S[i, j] = \frac{e^i_{sem} \cdot e^j_{sem}}
    {\|e^i_{sem}\| \|e^j_{sem}\|}
    \label{eq:semantic_similarity}
\end{equation}

with $e^i_{sem}, e^j_{sem} \in \mathbb{R}^d$ being the projected semantic embeddings of the $i$th and $j$th items in the sequence and $\|\cdot\|$ being the $\ell_2$ norm. The matrix of semantic similarities $\mathbf{S}$ captures semantic relatedness between all pairs of items and is used to bias the attention scores in the modified self-attention.

\subsubsection{Modified Self-Attention}
The scaled dot-product self-attention mechanism is based purely on learned query and key projections. Our proposed SISA attention mechanism adds the semantic similarity matrix $\mathbf{S}$ to the attention score matrix, enabling the semantic item relationships to be factored into the attention. In the input sequence $X = \mathbb{R}^{L \times d}$, the query ($Q$), key ($K$), and value ($V$) matrices are defined as in Vaswani et al.~\cite{ashish2017attention} as:

\begin{equation}
    \mathbf{Q} = XW_Q, \quad 
    \mathbf{K} = XW_K, \quad 
    \mathbf{V} = XW_V
    \label{eq:qkv}
\end{equation}

with $W_Q, W_K, W_V \in \mathbb{R}^{d \times d}$ being learnable projection matrices.The attention score matrix is then calculated following our proposal, which extends the original scaled 
bias term (see Vaswani et al.~\cite{ashish2017attention} for a description of the dot-product 
bias term:

\begin{equation}
    \mathbf{A} = \text{softmax}\left(\frac{\mathbf{Q}
    \mathbf{K}^\top}{\sqrt{d}} + \alpha \cdot 
    \mathbf{S}\right) \cdot \mathbf{V}
    \label{eq:modified_attention}
\end{equation}

where $d$ is the scaling factor to prevent the dot products from becoming too large, 
$\mathbf{S}$ is the semantic similarity matrix and $\alpha$ is a learnable scalar value controlling the weight of the semantic similarity in the attention scores. The use of $\alpha \cdot 
\mathbf{S}$ to the attention scores allows the model to pay more attention to semantically similar items, on top of the learned query-key interactions. The attention scores are masked to stop the model from tending to future items, maintaining the auto-regressive nature of sequential recommendations.

\subsubsection{Feed-Forward Network}
After the modified self-attention sub-layer, each position is fed into a position-wise feed-forward 
position separately, as proposed by Vaswani et al.~\cite{ashish2017attention}. 
This network is composed of two linear transformations with a GELU activation function~\cite{hendrycks2016gaussian}:

\begin{equation}
    \text{FFN}(h) = W_2 \cdot \text{GELU}(W_1 h + b_1) 
    + b_2
    \label{eq:ffn}
\end{equation}

where $W_1 \in \mathbb{R}^{d \times d_{ff}}$ and $W_2 \in \mathbb{R}^{d_{ff} \times d}$ are learnable weight matrices, $b_1 \in \mathbb{R}^{d_{ff}}$ and $b_2 \in \mathbb{R}^d$ are bias vectors, and $d_{ff}$ is the hidden size of the feed-forward network ($256$ in this work).

\subsubsection{Layer Normalization and Residual Connections}

Each sub-layer of the transformer encoder is followed by a layer normalization~\cite{ba2016layer} and residual connection~\cite{he2016deep}. In other words, for sub-layer function $\mathcal{F}(\cdot)$, the output is:

\begin{equation}
    h = \text{LayerNorm}(x + \mathcal{F}(x))
    \label{eq:layernorm_residual}
\end{equation}

where $x$ is the input to the sub-layer and $\mathcal{F}(x)$ is the output of the sub-layer. The residual connection alleviates the vanishing gradient problem during training~\cite{he2016deep}, and layer normalization helps the training process by normalizing the activations across the feature
dimension~\cite{ba2016layer}. The final layer of the transformer produces the contextual 
representations $H \in \mathbb{R}^{T \times d}$ that capture both behavioral and semantic information.

\subsection{Attention-Based Preference Aggregation}
The third novelty of the proposed SISA-Rec is the attention-based preference aggregation module. After the SISA transformer encoder, the contextual item representations $H \in \mathbb{R}^{T \times d}$ encode the sequential behavior patterns and semantic associations for each position in the user interaction. Instead of simply taking the last sequential state as the user representation, as is commonly done in this field~\cite{kang2018self}, the proposed aggregation module assigns each position a weight and then sums over the entire sequence, enabling the model to learn the contribution of each historical user interaction to the user's preference.

This work uses a learnable attention vector $w \in \mathbb{R}^d$ to calculate a scalar attention score for each potential position $t \in \{1, 2, \ldots, T\}$:

\begin{equation}
    a_t = w^\top h_t
    \label{eq:attention_score}
\end{equation}

with $h_t \in \mathbb{R}^d$ being the contextual representation at time $t$. The attention scores are then normalized over all non-padding positions via the softmax function to yield attention weights, as introduced in this paper:

\begin{equation}
    \alpha_t = \frac{\exp(a_t)}{\sum_{j=1}^{T} \exp(a_j)}
    \label{eq:attention_weights}
\end{equation}

with $\alpha_t$ being the attention weight at position $t$, after normalisation. The user preference vector $u \in \mathbb{R}^d$ is then defined as a linear combination of all contextual representations, as is proposed in this paper:

\begin{equation}
    u = \sum_{t=1}^{T} \alpha_t \cdot h_t
    \label{eq:user_preference}
\end{equation}

The user preference vector $u$ offers a comprehensive overview of the user's preferences, aggregating the contributions of all past interactions with weights learned from their importance. This approach is theoretically justified by the fact that user preferences are influenced by multiple past interactions, and not just the most recent one~\cite{kang2018self}, and that different interactions contribute differently to the description of current user preferences. The user preference vector $u$ is then fed to the scoring layer for item ranking.

\subsection{Scoring and Ranking}
\label{sec:scoring}
The scoring and ranking module calculates the relevance scores between the user preference vector $u$ and all candidate items, and returns the Top-N list of ranked items. The user preference vector $u \in \mathbb{R}^d$ from the preference aggregation module is used to compute the relevance score for a candidate item $i \in \mathcal{I}$ by taking the inner product with the fused item representation $v_i \in \mathbb{R}^d$ as in conventional inner product scoring for sequential recommendation~\cite{kang2018self}:

\begin{equation}
    s_i = u^\top v_i
    \label{eq:scoring}
\end{equation}

where $s_i \in \mathbb{R}$ is the predicted score for item $i$. The higher the score, the greater the predicted user-item affinity. The item representation $v_i$ is used for scoring the items instead of the original item identifiers to ensure that the semantic features of items are taken into account for ranking.

The Top-N recommendation list is generated by ranking all the candidate items in descending order of predicted relevance scores, and taking the top $N$ items according to the Top-N recommendation evaluation protocol of He et al.~\cite{he2017neural}:

\begin{equation}
    \hat{y} = \text{Top}N(\{s_i\}_{i \in \mathcal{I}})
    \label{eq:topn}
\end{equation}

with $\hat{y}$ being the list of $N$ items. In the evaluation protocol of this study, the target item is compared against 99 randomly selected negative items that the user has not interacted with, following the common sampled evaluation protocol in sequential recommendation~\cite{he2017neural}.

\section{Model Training and Validation}

The SISA-Rec model is trained in an end-to-end manner using a composite loss function that is a weighted sum of three terms: a Bayesian Personalized Ranking (BPR) loss, a contrastive alignment loss, and an $\ell_2$ regularization term.

\subsection {Bayesian Personalized Ranking Loss}

The main training criterion is derived from the Bayesian personalized ranking~\cite{rendle2012bpr}, which trains the model to give higher relevance scores for positive items (interacted with) than negative items (not interacted with). For every user $u \in \mathcal{U}$, a positive item $i^+ \in \mathcal{I}$ from the user's positive interaction history and a negative item $i^- \in \mathcal{I}$ from items $u$ has not interacted with are randomly sampled. The BPR loss is defined as \cite{rendle2012bpr}:

\begin{equation}
    \mathcal{L}_{BPR} = -\frac{1}{|\mathcal{U}|}
    \sum_{u \in \mathcal{U}} \log \sigma(s_{i^+} - s_{i^-})
    \label{eq:bpr}
\end{equation}

with $s_{i^+}$ and $s_{i^-}$ being the scores for the positive and negative items respectively, and $\sigma(\cdot)$ being the sigmoid function. The goal of learning is to minimize $\mathcal{L}_{BPR}$, which will ensure the model will consistently prefer observed interactions over unobserved ones.

\subsection {Contrastive Alignment Loss}

The contrastive alignment loss represents the fourth main contribution of the proposed SISA-Rec. It aims to promote consistency between the behavioral user representation $u \in \mathbb{R}^d$ generated by the preference aggregation module and a semantic user representation $u_{sem} \in \mathbb{R}^d$ derived from the average pooling of the BERT embeddings of the items in the user's interaction history, as per the semantic user representation approach of Hou et al.~\cite{10.1145/3534678.3539381}:

\begin{equation}
    u_{sem} = \frac{1}{T} \sum_{t=1}^{T} e^t_{sem}
    \label{eq:semantic_user}
\end{equation}

with $e^t_{sem} $ being the semantic projection of the item at position $t$ in the user's interaction history.

The contrastive loss is based on the InfoNCE objective~\cite{oord2018representation} that maximises the similarity between $u$ and $u_{sem}$ for the same user, and minimises similarity with semantic representations of other users in the batch. As proposed in this work, the InfoNCE objective is reformulated for aligning users' behavior and semantic representation in sequential recommendation:

\begin{equation}
    \mathcal{L}_{CL} = -\log \frac{\exp(u \cdot 
    u_{sem} / \tau)}{\sum_{j \neq i} \exp(u \cdot 
    u^j_{sem} / \tau)}
    \label{eq:contrastive}
\end{equation}

with $\tau$ a temperature hyperparameter to control the softness of the contrastive distribution, and $u^j_{sem}$ the semantic user representation of a different user $j$ in the batch. This loss function $\mathcal{L}_{CL}$ aims to align the behavioral interactions with the semantic meaning, so that the learned user representations become more semantically grounded.

\subsection {Regularization}

The loss function used when training the model is augmented with an $\ell_2$ regularization term to prevent overfitting by discouraging large parameter values, like typically in deep learning~\cite{kang2018self}:

\begin{equation}
    \mathcal{L}_{reg} = \|\Theta\|^2
    \label{eq:regularization}
\end{equation}

where $\Theta$ represents the model's parameters.

\subsection {Total Training Objective}

Following the proposed approach, the total loss (during the training) is defined as a weighted sum of the three loss components:

\begin{equation}
    \mathcal{L} = \mathcal{L}_{BPR} + \lambda_1 
    \mathcal{L}_{CL} + \lambda_2 \|\Theta\|^2
    \label{eq:total_loss}
\end{equation}

Here, $\lambda_1$ and $\lambda_2$ are hyperparameters that weight the loss functions and regularization term in the loss. In this study, $\lambda_1 = 0.1$ and $\lambda_2 = 1\times 10^{-4}$.

\subsection {Optimization and Validation}
The Adam optimizer is used to train the model with an initial learning rate of $1 \times 10^{-3}$ and a batch size of 256. A learning rate scheduler with a factor of 0.5 and patience of 2 epochs is used for reducing the learning rate if the validation performance stagnates. The model is trained for up to 30 epochs with early stopping based on the validation NDCG@10 metric, with a patience of 5 epochs. The best-performing model checkpoint, in terms of the highest validation NDCG@10, is used for testing.
The hyperparameter settings used in this study are listed in Table~\ref{tab:hyperparams}.
\begin{table}[ht]
\centering
\caption{Hyperparameter Configuration}
\label{tab:hyperparams}
\begin{tabular}{|l|c|}
\hline
\textbf{Parameter} & \textbf{Value} \\
\hline
Hidden dimensionality $d$ & 128 \\
\hline
Number of transformer layers $N$ & 2 \\
\hline
Number of attention heads & 2 \\
\hline
Feed-forward dimensionality $d_{ff}$ & 256 \\
\hline
Maximum sequence length $L$ & 50 \\
\hline
Dropout rate & 0.2 \\
\hline
Learning rate & $1 \times 10^{-3}$ \\
\hline
Batch size & 256 \\
\hline
Temperature $\tau$ & 0.1 \\
\hline
$\lambda_1$ & 0.1 \\
\hline
$\lambda_2$ & $1 \times 10^{-4}$ \\
\hline
Maximum epochs & 30 \\
\hline
Early stopping patience & 5 \\
\hline
\end{tabular}
\end{table}

\subsection{Evaluation Metrics}

We assess the performance of our proposed SISA-Rec model using two common ranking metrics for sequential recommendation: Hit Ratio (HR@$K$) and Normalized Discounted Cumulative Gain (NDCG@$K$) with $K \in \{5, 10\}$. These metrics are calculated on the Top-$N$ list of items recommended by the model and on the test set using the sampled evaluation protocol outlined in ~\ref{sec:scoring}.

\subsubsection{Hit Ratio}

Hit Ratio at $K$ (HR@$K$) is defined as the fraction of test cases where the ground truth target item appears in the Top $K$ items, as presented in He et al.~\cite{he2017neural}. Formally: 

\begin{equation}
\text{HR@}K = \frac{1}{|\mathcal{U}|} \sum_{u \in \mathcal{U}} \mathbb{1}[\text{rank}_u \leq K]
\label{eq:hr}
\end{equation}

with $\text{rank}_u$ being the rank of the ground truth item of user $u$ among the candidates and $\mathbb{1}[\cdot]$ being the indicator function that returns 1 if the condition is true and 0 otherwise. HR@$K$ is a binary metric, which does not consider the rank of the ground truth item in the Top-$K$ list, and is evaluated as the binary accuracy.

\subsubsection{Normalized Discounted Cumulative Gain}

Normalized Discounted Cumulative Gain at cutoff $K$, or NDCG@$K$ for short, is an extension of HR@$K$ that takes into account the rank of the target item in the list of recommendations and assigns higher values when the target item is ranked higher. The Discounted Cumulative Gain (DCG@$K$) for individual user $u$ is defined as in J\"{a}rvelin and Kek\"{a}l\"{a}inen~\cite{jarvelin2002cumulated} 
as:

\begin{equation}
\text{DCG@}K = \sum_{t=1}^{K} \frac{rel_t}{\log_2(t + 1)}
\label{eq:dcg}
\end{equation}

where the relevance $rel_t$ of the item at rank $t$ is 1 if the target item (the ground truth) is at rank $t$ and 0 otherwise. The discounting factor $\log_2(t + 1)$ down-weights the relevance of items that appear further down in the recommendation list, which reflects that users are more likely to view items ranked higher in the list.

As there is only one relevant item in each test case, the Ideal DCG (IDCG@$K$) is~\cite{jarvelin2002cumulated}:

\begin{equation}
\text{IDCG@}K = \frac{1}{\log_2(2)} = 1
\label{eq:idcg}
\end{equation}

The NDCG@$K$ for a user $u$ is then defined as~\cite{jarvelin2002cumulated}:

\begin{equation}
\text{NDCG@}K = \frac{\text{DCG@}K}{\text{IDCG@}K} = \frac{1}{\log_2(\text{rank}_u + 1)} \cdot \mathbb{1}[\text{rank}_u \leq K]
\label{eq:ndcg}
\end{equation}

The overall average NDCG@$K$ is given as~\cite{jarvelin2002cumulated}:

\begin{equation}
\overline{\text{NDCG@}K} = \frac{1}{|\mathcal{U}|} \sum_{u \in \mathcal{U}} \text{NDCG@}K_u
\label{eq:mean_ndcg}
\end{equation}

NDCG@$K$ offers a more detailed assessment of recommendation quality than HR@$K$ since it takes into account not only whether the target item is present in the Top-$K$ list, but also its position within the list. We report both of these metrics at $K = 5$ and $K = 10$ to give a well-rounded evaluation of recommendation quality across different list lengths.

\section{Results and Discussion}
\subsection{Experimental Setup}
The experiments were conducted on two Amazon product review datasets, Amazon Beauty and Amazon Toys \& Games, using a leave-one-out evaluation scheme, where the most recent interaction of every user was reserved for test and the second most recent for validation. During inference, the system was presented with the ground truth target item, along with 99 negative items, randomly sampled from the set of items with which the user has no interaction history. We measured the quality of recommender systems using the Hit Ratio (HR@$K$) and Normalized Discounted Cumulative Gain (NDCG@$K$) at $K \in \{5, 10\}$. Models were trained using the Adam optimiser with a fixed learning rate of $1 \times 10^{-3}$ and a batch size of $256$. Early stopping was used with a patience of five epochs to avoid overfitting, based on the NDCG@10 on a validation set.

We included four baselines to compare the performance of our model. PopRec is a non-personalised heuristic that sorts items based on the aggregate count of interactions in the training set, and provides a lower bound to the performance that quantifies the level of popularity bias inherent in the recommendation quality of the dataset. DuoRec~\cite{10.1145/3488560.3498433} is a contrastive learning-based sequential recommender that addresses representation degeneration through dropout-based augmentation and a uniformity regularization objective. SASRec~\cite{kang2018self} is a transformer-based sequential recommender that captures temporal patterns in interactions using causal self-attention applied only to item identifier embeddings, without access to item content or semantic information. BERT4Rec~\cite{sun2019bert4rec} is a bidirectional transformer-based sequential recommender that employs a masked item prediction objective to learn contextual user preference representations.

\subsection{Performance Comparison}

\begin{figure}[ht]
    \centering
    \includegraphics[width=1\textwidth]{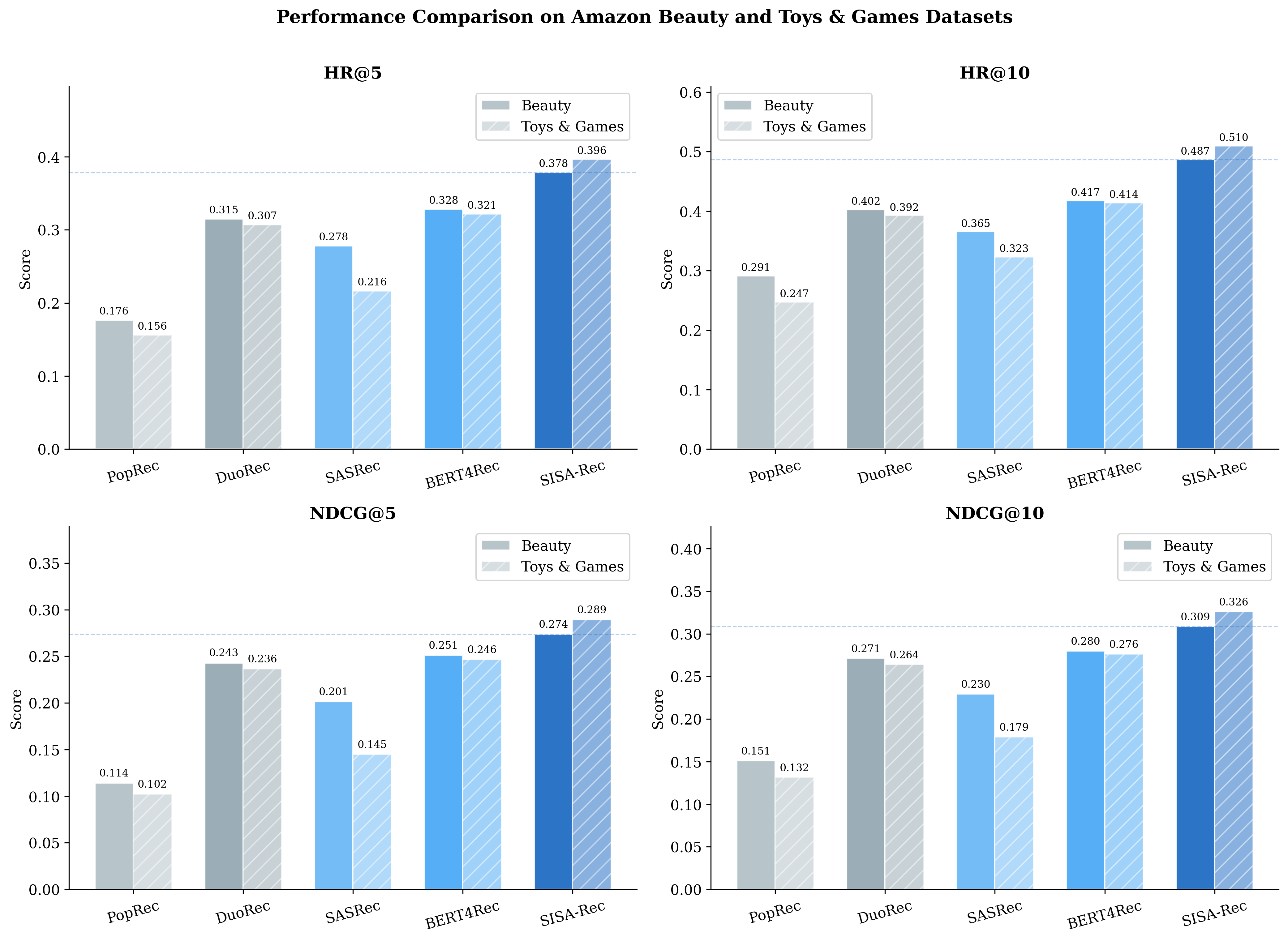}
    \caption{Performance comparison of all models on Amazon Beauty and Amazon Toys \& Games test sets across HR@5, HR@10, NDCG@5, and NDCG@10 metrics.}
    \label{fig:fig-4}
\end{figure}

The test-set results for all models on Amazon Beauty and Amazon Toys \& Games are presented in Table~\ref{tab:performance} and Fig.~\ref{fig:fig-4}. Across both datasets and all four metrics, SISA-Rec achieves the best performance, reaching an HR@10 of 0.4867 and NDCG@10 of 0.3087 on Amazon Beauty, and an HR@10 of 0.5097 and NDCG@10 of 0.3260 on Amazon Toys \& Games, while all trained baselines remain well above the PopRec popularity floor.

\begin{table}[t]
\centering
\caption{Performance comparison on Amazon Beauty and Amazon Toys \& Games test sets. Best results are in \textbf{bold}.}
\label{tab:performance}
\begin{tabular}{llcccc}
\hline
\textbf{Dataset} & \textbf{Model} & \textbf{HR@5} & \textbf{HR@10} & \textbf{NDCG@5} & \textbf{NDCG@10} \\
\hline
\multirow{5}{*}{Amazon Beauty}
 & PopRec   & 0.1765 & 0.2913 & 0.1141 & 0.1509 \\
 & DuoRec   & 0.3147 & 0.4024 & 0.2430 & 0.2712 \\
 & SASRec   & 0.2779 & 0.3654 & 0.2014 & 0.2296 \\
 & BERT4Rec & 0.3278 & 0.4174 & 0.2510 & 0.2799 \\
 & SISA-Rec & \textbf{0.3783} & \textbf{0.4867} & \textbf{0.2737} & \textbf{0.3087} \\
\hline
\multirow{5}{*}{Amazon Toys \& Games}
 & PopRec   & 0.1557 & 0.2472 & 0.1022 & 0.1315 \\
 & DuoRec   & 0.3067 & 0.3924 & 0.2365 & 0.2641 \\
 & SASRec   & 0.2162 & 0.3231 & 0.1447 & 0.1791 \\
 & BERT4Rec & 0.3210 & 0.4139 & 0.2464 & 0.2764 \\
 & SISA-Rec & \textbf{0.3962} & \textbf{0.5097} & \textbf{0.2893} & \textbf{0.3260} \\
\hline
\end{tabular}
\end{table}

\begin{figure}[ht]
\centering
\includegraphics[width=\linewidth]{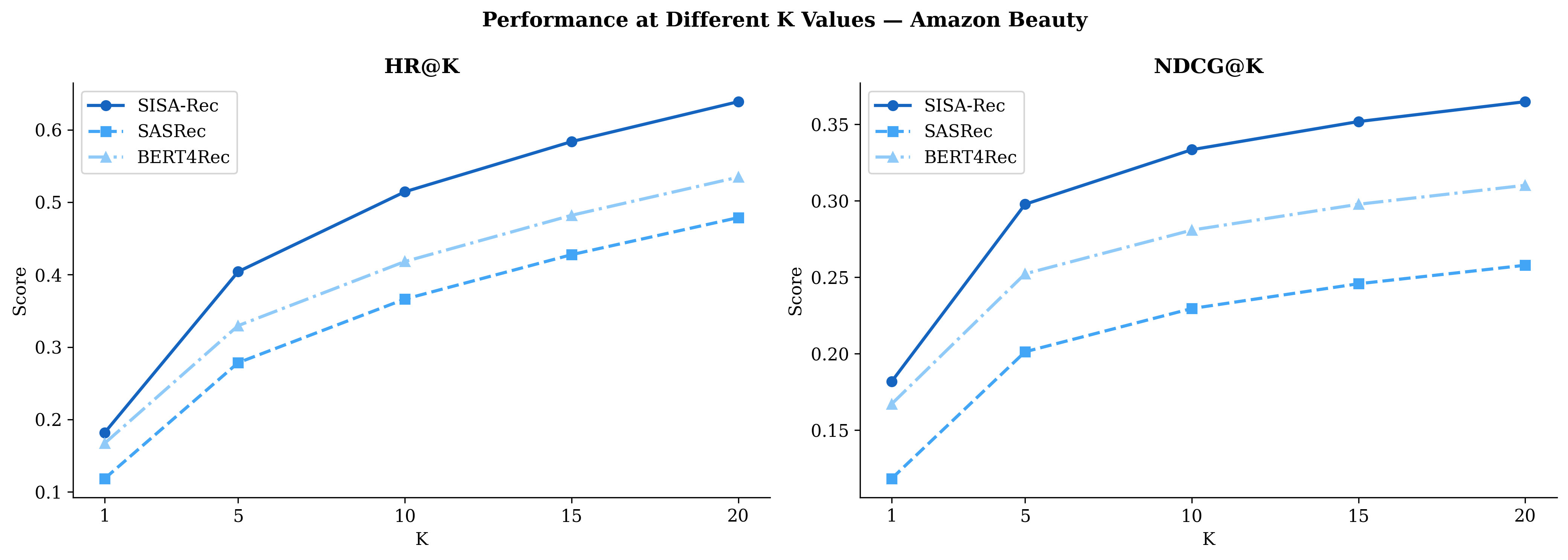}
\caption{Performance comparison of SISA-Rec, SASRec, and BERT4Rec at different cutoff values $K \in \{1, 5, 10, 15, 20\}$ on Amazon Beauty.}
\label{fig:k_values_beauty}
\end{figure}

\begin{figure}[ht]
\centering
\includegraphics[width=\linewidth]{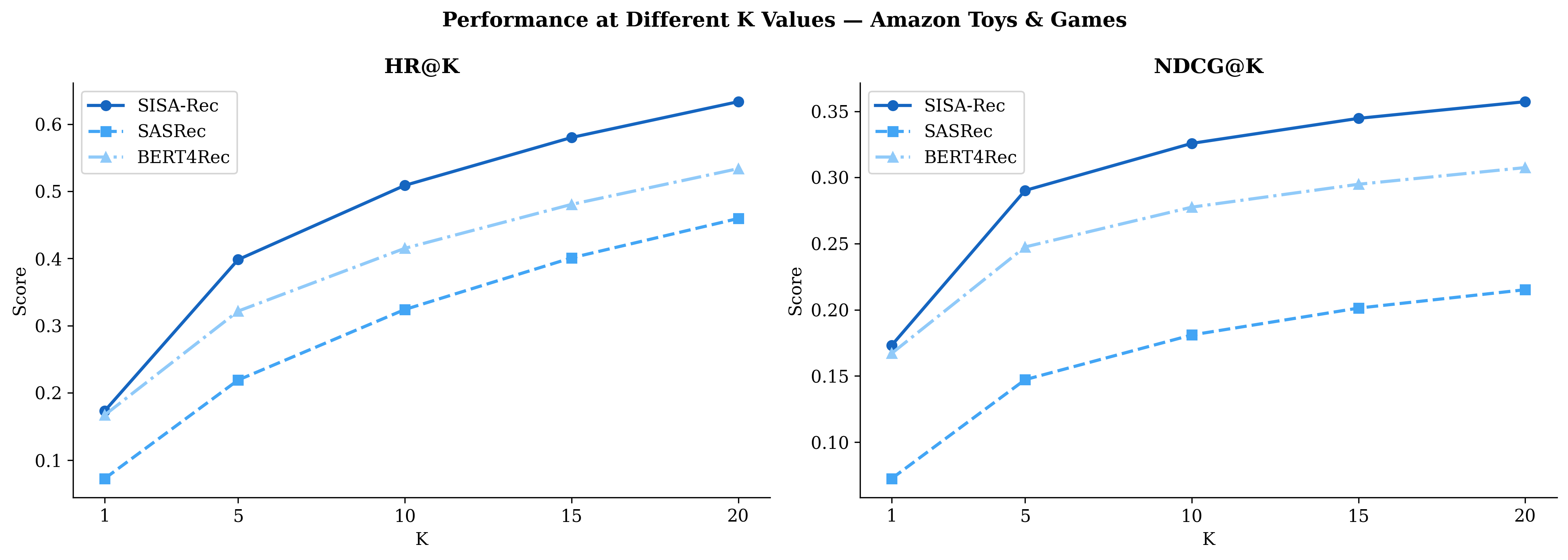}
\caption{Performance comparison of SISA-Rec, SASRec, and BERT4Rec at different cutoff values $K \in \{1, 5, 10, 15, 20\}$ on Amazon Toys \& Games.}
\label{fig:k_values_toys}
\end{figure}

All models were tested on the cut-off lengths $K \in \{1, 5, 10, 15, 20\}$ to further analyse the performance of the proposed model for varying lengths of the recommendation list. Fig.~\ref{fig:k_values_beauty} and Fig.~\ref{fig:k_values_toys} show the results. For both datasets, SISA-Rec shows a consistent improvement over all of the baselines, maintaining a clear margin across all cut-off lengths.

\subsection{Training Behaviour}
\begin{figure}[ht]
    \centering
    \includegraphics[width=1\textwidth]{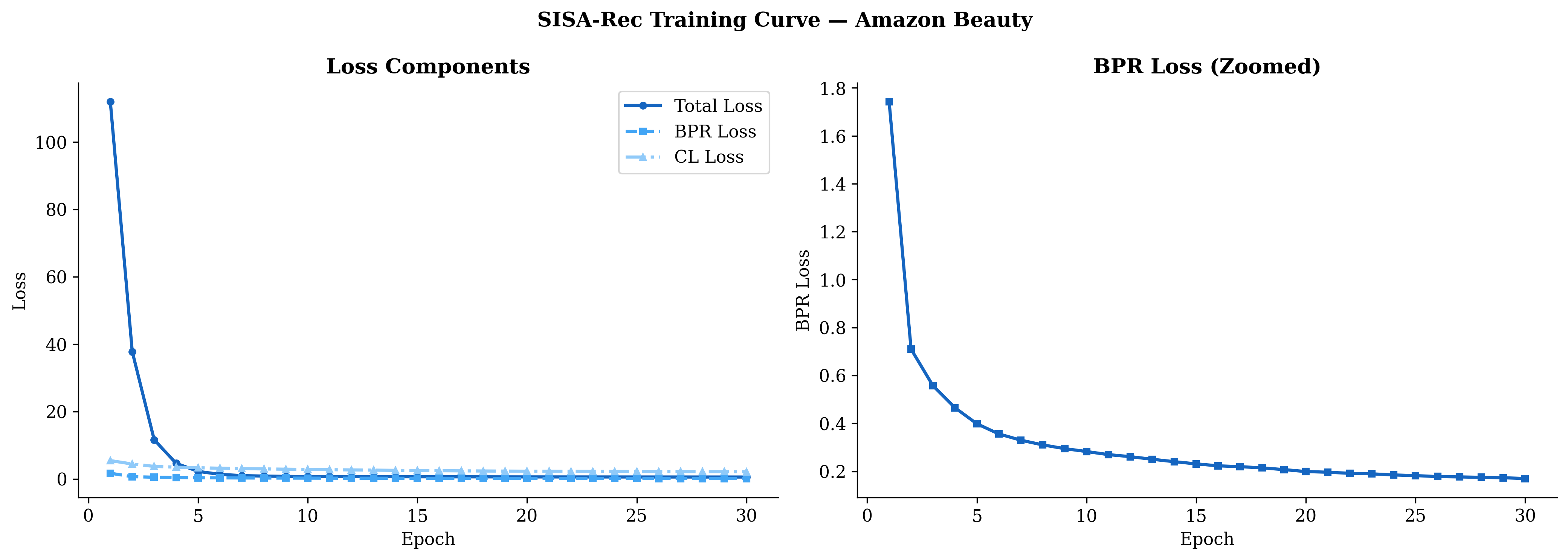}
    \caption{Training curve of SISA-Rec on Amazon Beauty: Total Loss, BPR Loss, and contrastive alignment loss per epoch.}
    \label{fig:training_beauty}
\end{figure}

\begin{figure}[ht]
    \centering
    \includegraphics[width=1\textwidth]{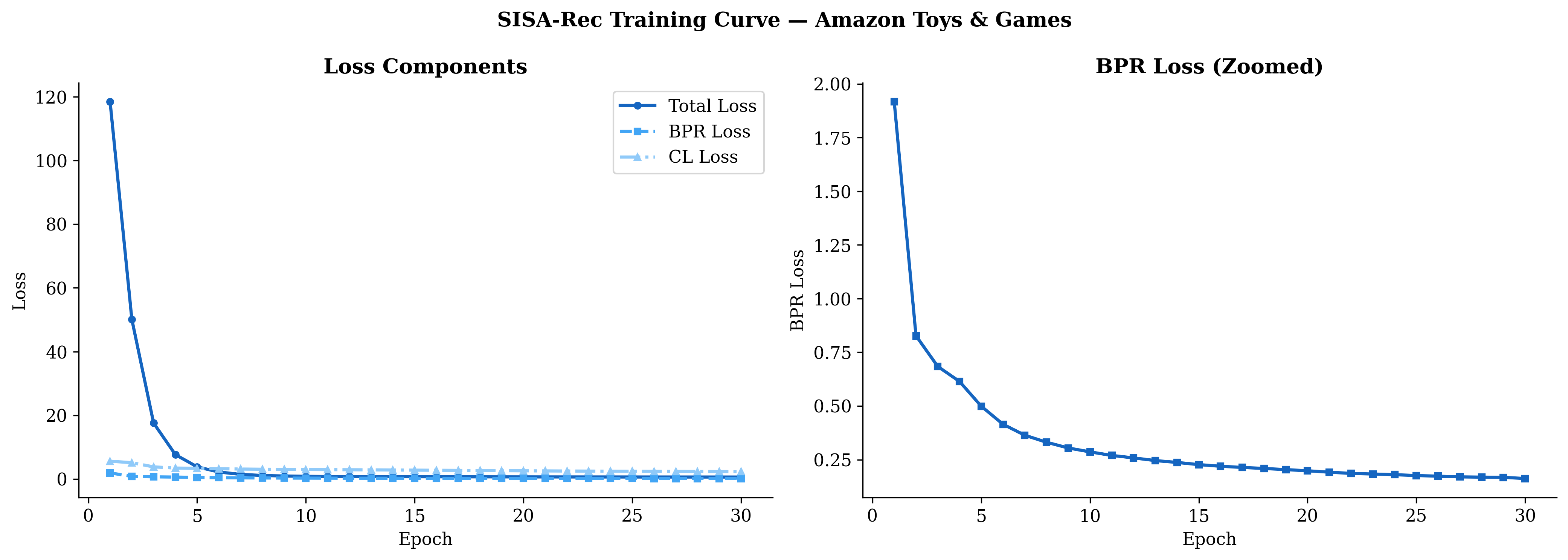}
    \caption{Loss, BPR loss, and contrastive alignment loss on SISA-Rec training curve on Amazon Toys \& Games.}
    \label{fig:training_toys}
\end{figure}

The training behaviour of SISA-Rec on the Amazon Beauty and Amazon Toys \& Games is depicted in Fig.~\ref{fig:training_beauty} and Fig.~\ref{fig:training_toys} respectively. In the first few epochs, the total loss drops rapidly for both datasets, then stabilizes in a smooth way over the course of one full training session (30 epochs) and is never observed to fluctuate or become unstable during training. The BPR loss is consistently decreasing over all the epochs, meaning that the model is getting better at learning to rank observed interactions higher than unobserved ones. The contrastive alignment loss is consistently low and does not increase with training, indicating that the behavioral and semantic user representations align early, stay consistent throughout training, and do not interfere with the main ranking objective. The objective of the joint training then yields stable and well-behaved optimization dynamics on both sparse Amazon datasets.

\subsection{Discussion}

SISA-Rec outperforms all baseline models across every evaluation metric on both
Amazon Beauty and Amazon Toys \& Games. The results show that models which
explicitly incorporate semantic item representations gain a measurable advantage
over those relying on behavioural co-occurrence alone, particularly on sparse
datasets where user preferences cannot be captured by interaction patterns by
themselves.

The gated fusion module allows the model to learn the relative contribution of
identifier-based and semantic information per dimension, rather than fixing it in advance. By injecting pairwise semantic similarity into the attention computation, the model attends more strongly to semantically related items when encoding a sequence, regardless of whether they co-occur in the observed history. The attention-based preference aggregation produces a more comprehensive user representation by drawing on all positions in the sequence rather than the final hidden state alone, which is especially valuable for the short interaction histories common in both datasets.

Relative to the strongest baseline, SISA-Rec improves HR@10 by 16.6\% and NDCG@10 by 10.3\% on Amazon Beauty, and by 23.1\% and 17.9\% respectively on Amazon Toys \& Games. The larger margin on Toys \& Games is consistent with its shorter average sequence length (8.63 interactions per user), which places greater demand on item-content signals when behavioural evidence is limited. The identity-only SASRec is the weakest of the trained models on both datasets, underlining the difficulty faced by purely identifier-based sequential models under high sparsity, where co-occurrence signal alone is insufficient.

\subsection{Cold-Start Analysis}

\begin{figure}
    \centering
    \includegraphics[width=0.7\textwidth]{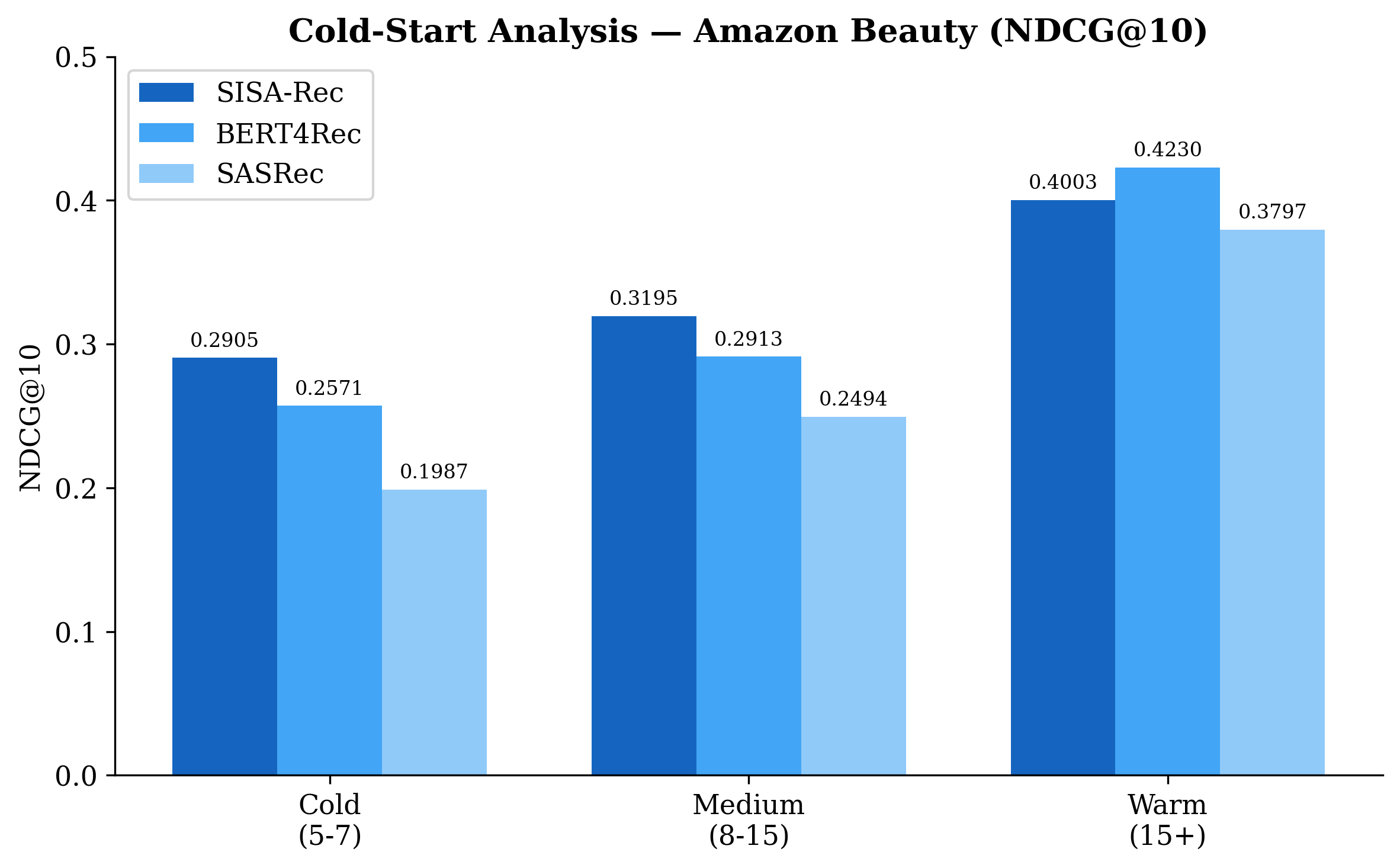}
    \caption{Cold-start analysis on Amazon Beauty: NDCG@10 for SISA-Rec, BERT4Rec, and SASRec across cold, medium, and warm user groups.}
    \label{fig:beauty_coldstart}
\end{figure}

\begin{figure}
    \centering
    \includegraphics[width=0.7\textwidth]{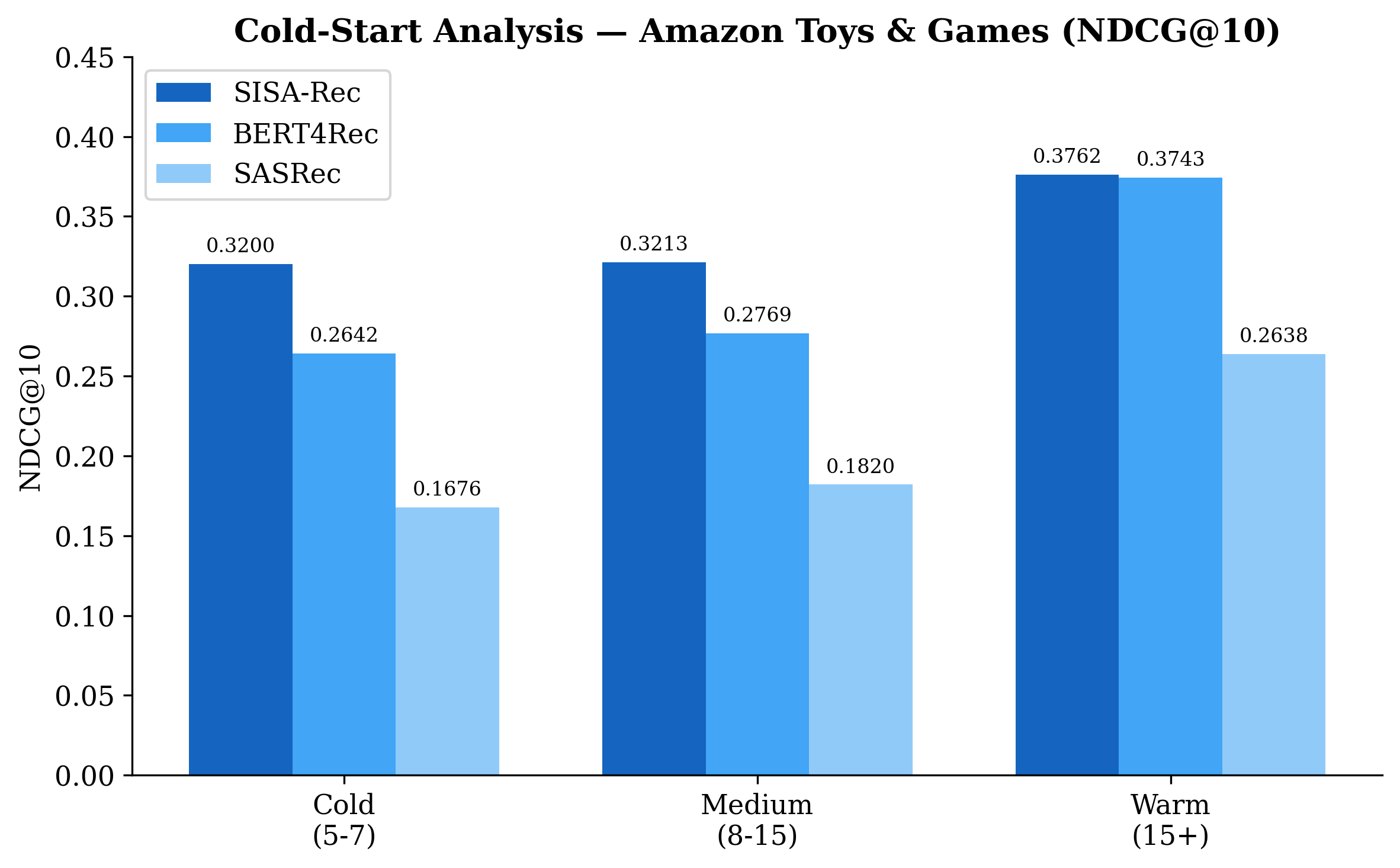}
    \caption{Cold-start analysis on Amazon Toys \& Games: NDCG@10 for SISA-Rec, BERT4Rec, and SASRec across cold, medium, and warm user groups.}
    \label{fig:toys_coldstart}
\end{figure}

The users were categorized into three groups on the basis of the number of sequence interactions: cold (5 to 7 interactions), medium (8 to 15 interactions), and warm (more than 15 interactions). Each group was calculated individually, and the results are shown in Table~\ref{tab:coldstart} and graphically in Fig.~\ref{fig:beauty_coldstart} and Fig.~\ref{fig:toys_coldstart} for SISA-Rec, BERT4Rec, and SASRec, respectively.

\begin{table}
\centering
\caption{Cold-start analysis on Amazon Beauty and Amazon Toys \& Games. NDCG@10 is reported for SISA-Rec, BERT4Rec, and SASRec across three user interaction-density groups.}
\label{tab:coldstart}
\begin{tabular}{llccc}
\hline
\textbf{Dataset} & \textbf{User Group} & \textbf{SISA-Rec} & \textbf{BERT4Rec} & \textbf{SASRec} \\
\hline
\multirow{3}{*}{Amazon Beauty}
 & Cold (5--7)    & \textbf{0.2905} & 0.2571 & 0.1987 \\
 & Medium (8--15) & \textbf{0.3195} & 0.2913 & 0.2494 \\
 & Warm (15+)     & 0.4003 & \textbf{0.4230} & 0.3797 \\
\hline
\multirow{3}{*}{Amazon Toys \& Games}
 & Cold (5--7)    & \textbf{0.3200} & 0.2642 & 0.1676 \\
 & Medium (8--15) & \textbf{0.3213} & 0.2769 & 0.1820 \\
 & Warm (15+)     & \textbf{0.3762} & 0.3743 & 0.2638 \\
\hline
\end{tabular}
\end{table}

The two datasets reveal a very clear and consistent pattern: the benefit of semantic integration is greatest when the behavioural signal is least. On Amazon Toys \& Games, SISA-Rec achieves an NDCG@10 of $0.3200$ in the cold group, well ahead of BERT4Rec ($0.2642$) and SASRec ($0.1676$). The lead also remains in the medium group ($0.3213$ compared to the models based on identity or content, which are $0.2769$ and $0.1820$ respectively); in the warm group, the gap becomes smaller because the longer history helps the models based on identity/content to close the performance gap.

Likewise, in the compressed version on Amazon Beauty. SISA-Rec leads clearly in the cold group (NDCG@10 of $0.2905$ versus $0.2571$ for BERT4Rec and $0.1987$ for SASRec) and in the medium group ($0.3195$ versus $0.2913$ and $0.2494$). The models, however, merge in the warm group, with BERT4Rec slightly outperforming SISA-Rec ($0.4230$ vs $0.4003$), as longer interaction history is sufficient enough to provide enough behavioural signal for the identity-based models to become competitive with the semantically grounded ones. These results illustrate that the semantic integration is most important for the interaction histories which are short and that SISA-Rec outperforms purely behavioural baselines most in cold start conditions and that the benefit of the semantic integration compared to the purely behavioural baselines is reduced as more interaction data is used.

\FloatBarrier
\clearpage

\section{Conclusion}

We present SISA-Rec, a semantically integrated sequential recommendation framework designed to address the limitations of transformer-based architectures in sparse and cold-start environments. Previous models that rely primarily on item ID embeddings, our approach merge textual context directly into the sequential architecture via a gated fusion module, semantic-aware self-attention, attention-based preference aggregation, and a joint optimization objective that aligns behavioral and semantic representations. Evaluation metrics results on the two highly sparse Amazon Beauty and Amazon Toys \& Games datasets shows that SISA-Rec outperforms strong baselines across all metrics. Compared to the top-performing baseline, the model improved by 16.6\% in HR@10 and 10.3\% in NDCG@10 on Amazon Beauty, whereas, on Amazon Toys \& Games dataset it improved 23.1\% in HR@10 and 17.9\% in NDCG@10. These results evident that the proposed framework better captures item relationships when behavioral records are limited. Future work will explore multimodal representations including images and user reviews, scaling the framework to larger datasets, and developing adaptive integration techniques to further optimize performance across various domains.
\FloatBarrier
\bibliographystyle{elsarticle-num}
\bibliography{cas-refs.bib}

\end{document}